%% file: main.tex
\documentclass[10pt,twocolumn,letterpaper]{article}

\usepackage{cvpr}              % To produce the CAMERA-READY version
\usepackage{algorithm}
\usepackage{algpseudocode}
\usepackage{algorithmicx}
\usepackage{multirow}

\input{preamble}

\definecolor{cvprblue}{rgb}{0.21,0.49,0.74}
\usepackage[pagebackref,breaklinks,colorlinks,citecolor=cvprblue]{hyperref}

\def\paperID{15925} % *** Enter the Paper ID here
\def\confName{CVPR}
\def\confYear{2024}

\title{Rethinking the Spatial Inconsistency in Classifier-Free Diffusion Guidance}

\author{Dazhong Shen$^1$,  ~~~ Guanglu Song$^2$, ~~~ Zeyue Xue$^3$, ~~~ Fu-Yun Wang$^4$, ~~~ Yu Liu$^{1,2,\thanks{the corresponding author: liuyuisanai@gmail.com}}$\\
$^1$Shanghai Artificial Intelligence Laboratory, ~~$^2$SenseTime Research, \\
$^3$The University of Hong Kong, ~~
$^4$The Chinese University of Hong Kong \\
}

\begin{document}
\maketitle
\input{sec/0_abstract}    
\input{sec/1_intro}
\input{sec/2_relatedwork}
\input{sec/3_preliminary}

\input{sec/4_methods}

\input{sec/5_experiment}
\input{sec/6_conclusion}

% \newpage
{
    \small
    \bibliographystyle{ieeenat_fullname}
    \bibliography{main}
}
\newpage
\input{sec/X_suppl}

\end{document}

%% file: preamble.tex
\usepackage[dvipsnames]{xcolor}

%% file: sec/0_abstract.tex
\begin{abstract}
Classifier-Free Guidance (CFG) has been widely used in text-to-image diffusion models, where the CFG scale is introduced to control the strength of text guidance on the whole image space.
However, we argue that a global CFG scale results in spatial inconsistency on varying semantic strengths and suboptimal image quality.
To address this problem, we present a novel approach, Semantic-aware Classifier-Free Guidance (S-CFG), to customize the guidance degrees for different semantic units in text-to-image diffusion models.
Specifically, we first design a training-free semantic segmentation method to partition the latent image into relatively independent semantic regions at each denoising step. 
In particular, the cross-attention map in the denoising U-net backbone is renormalized for assigning each patch to the corresponding token, while the self-attention map is used to complete the semantic regions.
Then, to balance the amplification of diverse semantic units, we adaptively adjust the CFG scales across different semantic regions to rescale the text guidance degrees into a uniform level. 
Finally, extensive experiments demonstrate the superiority of S-CFG over the original CFG strategy on various text-to-image diffusion models, without requiring any extra training cost. our codes are available at \url{https://github.com/SmilesDZgk/S-CFG}.

\end{abstract}

%使用 mean norm 
%展示 case attention map 

%% file: sec/1_intro.tex
\vspace{-2mm}
\section{Introduction} \label{sec:intro}
\vspace{-1mm}
%background
% With the advent of image-generative models and language models, 
Recently, text-to-image generation has witnessed rapid development and various applications~\cite{reed2016generative,xu2018attngan,ramesh2021zero,rombach2022high,ramesh2022hierarchical}, where visually stunning images can be created by simply typing in a text prompt.
In particular, after DDPM~\cite{ho2020denoising, dhariwal2021diffusion} succeeded GANs~\cite{goodfellow2020generative,brock2018large}, diffusion models~\cite{sohl2015deep}, such as Stable Diffusion~\cite{rombach2022high} and DallE-3~\cite{Betker2023Improving}, have emerged as the new state-of-the-art family for image-generative models.
% They excel at learning the data distribution by utilizing a noise/score prediction objective and iteratively removing noise from initial vectors during the sampling process.
% The key feature of diffusion models is to learn the image data distribution through a score prediction objective and progressively remove noise from the initial vectors in the iterative sampling stage.

The key feature of diffusion models is to approximate the true data distribution $p(x)$ by reversing the process of perturbing the data with noise progressively in a long iterative chain.
To incorporate the text prompt $c$ into the final generation, it is necessary to enhance the likelihood of $c$ given the current latent image $x_t$ at each reversed diffusion step $t$.
% As a result, to embed the text prompt $c$ in the final generation, we need to increase the likelihood of $c$ given the current latent image $x_t$ at each reversed diffusion step $t$.
Instead of training extra classifiers to model $p(c|x_t)$ at each diffusion step $t$~\cite{dhariwal2021diffusion},  classifier-free guidance (CFG)~\cite{ho2021classifier} has recently been proposed to estimate both the classifier score $\nabla_{x_t} \log p(c|x_t)$ and the diffusion score $\nabla_{x_t} p(x_t)$  with the same neural models, such as U-net~\cite{ronneberger2015u}.
In particular, an empirical CFG scale is introduced to control the strength of the text guidance on the whole image space.

% \begin{figure}[t]
%   \centering
%   \includegraphics[width=1.0\linewidth]{figs/case.jpg}
%   \caption{\textbf{A motivation example.} The left image is generated by Stable Diffusion with DDIM sampler, where the segmentation is manually labeled. The right shows the average norm curves of the estimated classifier score $\  \nabla_{x_t} \log p(c|x_t)$ (solid line) and diffusion score  $ 
% \nabla_{x_t} \log p(x_t)$ (dashed line) in each semantic region. The Y-axis scale unit is set as the dynamic variance parameter $\sigma_t$ for better illustrations without damaging the conclusion.}
%   \label{fig:example}
%   \vspace{-2mm}
% \end{figure}

\begin{figure}[t]
  \centering
  \includegraphics[width=0.95\linewidth]{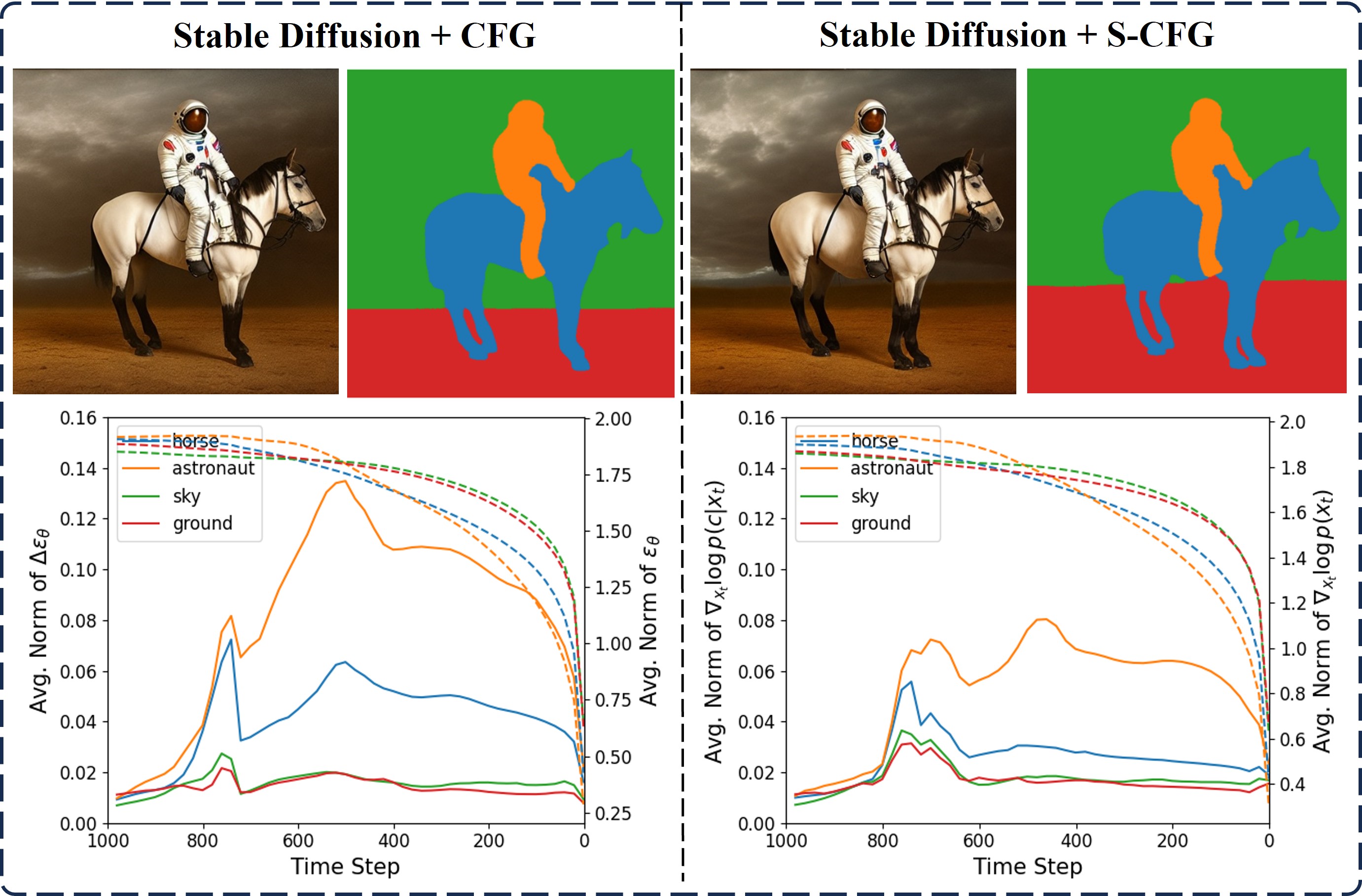}
  \caption{\textbf{A motivation example.} { The first line shows images generated by Stable Diffusion with  CFG  and  S-CFG, where the prompt is ``\textit{a photo of an astronaut riding a horse}'' and the segmentation maps are manually labeled (Ground, Sky, Horse, Astronaut). The below line shows the average norm curves of the estimated classifier score $\  \nabla_{x_t} \log p(c|x_t)$ (solid line) and diffusion score  $\nabla_{x_t} \log p(x_t)$ (dashed line) in each semantic region. The Y-axis scale unit is set as the dynamic variance parameter $\sigma_t$ for better illustrations without damaging the conclusion.}}
  
%   \caption{\textbf{A motivation example.} The left image is generated by Stable Diffusion with DDIM sampler, where the segmentation is manually labeled. The right shows the average norm curves of the estimated classifier score $\  \nabla_{x_t} \log p(c|x_t)$ (solid line) and diffusion score  $ 
% \nabla_{x_t} \log p(x_t)$ (dashed line) in each semantic region. The Y-axis scale unit is set as the dynamic variance parameter $\sigma_t$ for better illustrations without damaging the conclusion.}
  \label{fig:example}
  \vspace{-2mm}
\end{figure}

However, we argue that \textit{ a global CFG scale results in spatial inconsistency on varying semantic strengths during the denoising process and suboptimal quality of the final image.} 
%(see Figure~\ref{fig:example}).
% However, we argue that \textit{the ideal CFG scale setting should be spatial-aware} (see Figure~\ref{fig:example}), resulting in imbalanced amplification for different image semantic units at each denoising step and suboptimal sampling quality directly with a global CFG scale.
% For example, 
Figure~\ref{fig:example} shows samples generated by Stable Diffusion~\cite{rombach2022high}. The images can be segmented into four semantic regions corresponding to ``astronaut'', ``horse'', ``sky'' and ``ground''. 
To compare the guidance degrees assigned to different semantic units, the figures in the second line illustrate the average norm curves of the estimated classifier score $\nabla_{x_t} \log p(c|x_t)$ and diffusion score  $\nabla_{x_t} \log p(x_t)$ in each semantic region at any time step.  
% Based on the results in the right of Figure~\ref{fig:example}, 
{ as for the images with the original CFG strategy,} we can find that the classifier score norm {changes} a lot on different semantic units, while the norms of diffusion scores seem to be closer. 
Intuitively, the larger classifier score implies a greater guidance degree received by the semantic unit. 
As a result, the final generative samples may exhibit spatial inconsistency in image qualities for different semantic units. For instance, the {``astronaut''} region, which consistently attains the highest score ratio, displays intricate and finely detailed structures that starkly contrast with the ``sky'' and ``ground'' regions.
% In particular,  Even though both ``horse'' and ``astronaut'' are included in the prompt, there exists a significant disparity in their score ratio curves, which also leads to imbalanced levels of image details within their respective regions.

% To alleviate this imbalance, some researchers proposed to renormalize the latent image $x_t$ by setting the adaptive CFG scale on each patch~\cite{github2023mcmonkey}. However, this approach suffers an unstable image sampling quality due to the disruption of relative scale relations among patches, particularly within each semantic region, where mutual dependencies between different patches are significant.
Along this line, in contrast to the previous works, we propose to set customized CFG scales for different semantic regions of the latent image at each denoising step.
In particular, we assume that the inter-patches in each semantic region serve a similar semantic concept and different regions are relatively independent.
In this case, the classifier scores $\nabla_{x_t} \log p(c|x_t)$ can be approximately deduced into the combination of that conditioning on all independent semantic regions. 
Therefore, customized CFG scales can be safely involved for each semantic region, without the disruption of relative relations among interdependent patches.
However, it is not trivial to conduct semantic segmentation on the latent image without accessing the final generated image. 
% Complex computation should also be avoided for efficiency.
Meanwhile, determining the customized CFG scales to balance semantic units is another challenge.

To this end, in this paper, we propose a novel approach, called Semantic-aware Classifier-Free Guidance (S-CFG), to dynamically and customizedly control the text guidance degrees in text-to-image diffusion models.
Specifically, when modeling the conditional distribution $p(x|c)$, diffusion models take $c$ as another input with self-attention and cross-attention layers to mix up the image and text, which preserves the underlying semantic information.
Along this line, we first design a training-free segmentation method for the latent images at each denoising step.
In particular, the cross-attention map in the denoising U-net backbone is renormalized for assigning each patch to the corresponding token, while the self-attention map is used to complete the semantic regions.
% Along this line, we first design an efficient attention-based method to segment the latent image at each denoising step by assigning the patch to the token.
% In particular,  this method is efficient and robust without the need for any additional learnable modules.
Then, to balance the amplification of diverse semantic information, we rescale the classifier score $\nabla_{x_t} \log p(c|x_t)$ across different semantic regions to a uniform level with the adaptive CFG scales.
% scale ratio with respect to the diffusion score $\nabla_{x_t} \log p(x_t)$.
% where the adaptive CFG scales are set proportional to that ratio between average norms of the whole image and the current semantic region.
Finally, we conduct qualitative and quantitative analysis based on various diffusion models. The results demonstrate that S-CFG can outperform the original CFG strategy and obtain a robust improvement without any extra training cost{.}
{ 
At first glance, the right part in Figure~\ref{fig:example} demonstrates reduced disparities among the classifier score norms $\nabla_{x_t} \log p(c|x_t)$  of different semantic units in the image with S-CFG. As a result, more abundant clouds float in the ``sky''. The boundary between the ``sky'' and the ``ground'' is clearer.
% The two front hooves of the ``horse'' are easier to distinguish.
% The right part of Figure~\ref{fig:example} demonstrates reduced disparities among the classifier score norms $\nabla_{x_t} \log p(c|x_t)$ for different semantic units with S-CFG. Consequently, a greater number of clouds are visible in the ``sky'', and the boundary between the ``sky'' and the ``ground'' is more distinct. Moreover, the distinction between the two front hooves of the ``horse'' is enhanced.
}

%% file: sec/2_relatedwork.tex
\vspace{-1mm}
\section{Related Work}
\label{sec:formatting}
\vspace{-2mm}

\subsection{Image Diffusion Generative Models}
\vspace{-2mm}
%background
% With the advantage of deep learning, a wide range of generative models have emerged to produce images of exceptional quality. Early works in this field focused on the context of GANs~\cite{goodfellow2020generative,brock2018large} and VAEs~\cite{kingma2013auto,razavi2019generating}, which all build a mapping from a simple distribution, such as Gaussian distribution, to the real and complex image data distribution in a one-shot manner. 
%diffusion model
Recently, diffusion models have emerged as an expressive and flexible family for image generation with remarkable image quality and various applications~\cite{ramesh2021zero,rombach2022high,ramesh2022hierarchical, bar2022text2live,kim2022diffusionclip,ho2022cascaded,lugmayr2022repaint}.
The general idea is to apply a forward diffusion process that adds tiny noise to the input data, then learn the reverse process with neural networks to gradually recover the original samples from the noisy data, step-by-step. 
Among them, Denoising Diffusion Probabilistic Model (DDPM)~\cite{ho2020denoising} is the representative baseline, which carefully designed the noise schedule on the pixel space during the forward process and the network architecture in the reverse process. As a result, diffusion models achieved better model coverage and training stability compared to GANs~\cite{goodfellow2020generative,brock2018large,karras2020analyzing}.
To further reduce {computational} costs, the subsequent study turned to combining DDPM and VAE~\cite{kingma2013auto,razavi2019generating,ijcai2021p408} by applying diffusion models to the lower-dimensional latent space of {a} VAE trained on large-scale image datasets, such as Stable Diffusion~\cite{rombach2022high}. 
% Along this line, diffusion models can generate higher-resolution images, even for 1K resolution~\cite{}. 
In general, diffusion models suffer the downside of low inference speed compared to other generative models. However, this problem can be greatly alleviated by distillation strategies~\cite{song2023consistency,wang2024animatelcm} or advanced sampling strategies, such as DDIM~\cite{song2020denoising,zhang2022gddim}, {DPMSolver}~\cite{lu2022dpm1,lu2022dpm2}, PNDM~\cite{karras2022elucidating}, {Euler}~\cite{karras2022elucidating}, and {DEIS~\cite{zhang2022fast}},  which can perform 10X to 100X speedup compared to the original DDPM sampler. 
Here, we further explore {a better} way for image generation based on diffusion models.

\vspace{-1mm}
\subsection{Text-guided Generation}
\vspace{-2mm}
% Similar to other generative models~\cite{goodfellow2020generative,brock2018large,kingma2013auto,razavi2019generating}, diffusion models are capable of modeling conditional information during image generation, such as class labels, semantic maps, etc. In particular, 
Recently, the text-guided generation in diffusion models has reached an unprecedented level, like DallE-3~\cite{Betker2023Improving}.
This generative power stems from three aspects. 
First, to represent the unstructured text, expressive language embedding models are used to embed each token in the given text, such as CLIP~\cite{radford2021learning} in Stable Diffusion~\cite{rombach2022high}, and T5~\cite{raffel2020exploring} in Imagen~\cite{saharia2022photorealistic}.
Second, to facilitate the interaction between text and image information, diffusion models typically enhance the network backbone, such as the U-net backbone~\cite{ronneberger2015u}, with the cross-attention mechanism.
This mechanism involves utilizing the image embedding as the query and the key and value embeddings derived from the text.
Third,  Classifier-Free Guidance (CFG)~\cite{ho2021classifier}  has recently been widely involved as a lightweight and robust technique to encourage text prompt adherence in generations. 
Instead of training extra classifiers~\cite{dhariwal2021diffusion,liu2023more}, CFG mixes the score estimates of the diffusion model with or without the conditional prompt. Some other works~\cite{liu2022compositional,huang2023composer} further separate a prompt into multiple concepts and generate an image by combining a set of diffusion models with each of them conditioning on a certain concept component.
% They all use the global scales to control the guidance degree of different conditions on the whole image. }
Here, we further emphasize the importance of varying CFG scales across different image semantic regions and design the semantic-ware CFG strategy to improve image quality.

{
\vspace{-1mm}
\subsection{Applications with Cross-Attention Maps}
\vspace{-2mm}
Cross-attention maps in the diffusion U-net Backbone are derived to represent the spatial relation between image patches and prompt tokens. They provide valuable semantic information for image segmentation and can contribute to various applications. For example, some works~\cite{couairon2023zero,chen2024training,zhao2023loco,xie2023boxdiff} introduce layout control in image generation by minimizing the difference between the cross-attention-based semantic segmentation and the given layout conditions. Prompt2Prompt~\cite{hertz2022prompt} achieves image editing by simply replacing, adding, or re-weighting cross-attention maps.  
Attend-and-Excite~\cite{chefer2023attend} improves the text alignment by optimizing the cross-attention maps during the inference process.
Subsequent works further extend those ideas for image-to-image translation~\cite{parmar2023zero}, text-driven image editing~\cite{wang2023dynamic,guo2023focus}, and compositional image generation~\cite{wang2023compositional}.
In this paper, we further use cross-attention maps to improve image quality by segmenting latent images and customizing the guidance degrees of different semantic regions.
}
% Moreover, CFG has the potential to be applied to any natural text prompts in principle, not only for class labels, making it a versatile and adaptable method~\cite{}.

%Classifier guidance 
%Classifier-free guidance CLIP 

%% file: sec/3_preliminary.tex
\vspace{-1mm}
\section{Preliminary}
\vspace{-2mm}

\subsection{Diffusion Models}
\vspace{-2mm}
Given the image data space $\mathcal{X}$, diffusion models define a Markov Chain, known as the forward process, to corrupt the real data $x_0 \in \mathcal{X}$ by progressively adding Gaussian noise from time steps $0$ to $T$:
\begin{equation}
    \begin{split}
        q(x_t|x_{t-1}) = \mathcal{N}(x_t; \sqrt{1-\beta_t}x_{t-1}, \beta_t \textbf{I}),
    \end{split}
\end{equation}
where $\{\beta_t\}_{t=1:T}$ denotes the variance for each noise step, set as constant usually. 
Taking advantage of the properties of the Gaussian distribution, we can obtain $x_t$ at an arbitrary time step $t$ using the following closed form:
% In order to obtain $x_t$ at an arbitrary time step $t$ without $t$ steps iterations, a parameterization trick enabled by the properties of the Gaussian distribution might be performed:
\begin{equation}
    \begin{split}
        % q(x_t|x_0) = \mathcal{N}(x_t; \sqrt{\overline{\alpha}_t} x_0, (1-\overline{\alpha}_t)\textbf{I}),\\
        x_t =\sqrt{\overline{\alpha}_t} x_0+\sqrt{1-\overline{\alpha}_t}\epsilon_t,~\epsilon_t \sim \mathcal{N}(0, \textbf{I}),
    \end{split}\label{equ:xtx0}
\end{equation}
where $\alpha_t = 1-\beta_t$ and $\overline{\alpha}_t = \prod_{s=1}^t \alpha_s$. $x_T$ will degrade to standard Gaussian noise with $\overline{\alpha}_T \approx 0$.

The reverse denoising process aims to approximate the true posterior of each forward step via a time-dependent neural network parameterized by $\theta$:
\begin{equation}
 \begin{split}
     p_{\theta} (x_{t-1}|x_t) = \mathcal{N}(x_{t-1}; \mu_{\theta}(x_t, t),\sigma_{\theta}(x_t,t) \textbf{I}),
 \end{split}
\end{equation}
which can be used to generate image $x_0 \sim p_{\theta} (x_0)$ by sampling Gaussian noise $x_T \sim \mathcal{N}(0, \textbf{I})$ first and denoising {step-by-step} from $x_{T-1}$ to $x_0$.
In practice, to {simplify} the model training, $\sigma_{\theta}(x_t,t)$ is set as constant $\sigma_t$~\cite{dhariwal2021diffusion} and $\mu_{\theta}(x_t, t)$ is parameterized as follows:
\begin{equation}
    \begin{split}
        \mu_{\theta}(x_{t}, t) = \frac{1}{\sqrt{\alpha_t}} \left (x_t- \frac{\beta_t}{1-\overline{\alpha}_t} \epsilon_{\theta}(x_t, t)\right ),
    \end{split}
\end{equation}
where the neural model $\epsilon_{\theta}$, such as U-net~\cite{ronneberger2015u}, is trained to predict the noise $\epsilon_t$ added in each forward step, which also mirrors the denoising score-matching, i.e, $\epsilon_{\theta}(x_t,t) \approx -\sigma_t\nabla_{x_t} \log p(x_t)$.

% The simplified training objective is derived as in :
% \begin{equation}
%     \begin{split}
%         \mathcal{L} = \mathbb{E}_{t\sim [1,T], x_0\sum q(x), \epsilon \sim \mathcal{N}(0, I)} [||\epsilon - \epsilon _t(x_t, t)||^2]
%     \end{split}
% \end{equation}

% epsilon vs score matching 

\vspace{-1mm}
\subsection{Classifier-free Guidance}
\vspace{-2mm}
%Stay on topic with Classifier-Free Guidance
%CLASSIFIER-FREE DIFFUSION GUIDANCE
% analysis with simple Gaussian 
The vanilla diffusion model described above is an unconditional generative model $p_{\theta}(x_0)$  to approximate the true data distribution $q(x_0)$. However, in {practical} scenarios, there is a growing demand to condition the generation on a label or text prompt $c$~\cite{zhang2023text}. 
To address this requirement, 
classifier-guidance~\cite{dhariwal2021diffusion} incorporates an auxiliary classifier $p_{\phi}(c|x_t)$ to guide the sampling in each reverse denoising step, thereby increasing the likelihood of $c$ given $x_t$. Specifically, the diffusion score is modified as follows:
\begin{equation}
    \begin{split}
        \hat{\epsilon}_{\theta}(x_t. c, t) = \epsilon_{\theta}(x_t, t)-\gamma \sigma_t \nabla_{x_t}  \log p_{\phi}(c|x_t) \\
        \approx -\sigma_t \nabla_{x_t} \log (p(x_{t})p^{\gamma}_{\phi}(c|x_t)),
    \end{split}\label{equ:cg}
\end{equation}
where $\gamma$ is a scalar parameter to regulate the strength of the classifier guidance.
While this method has demonstrated some performance improvements, training {a} robust classifier for all reverse steps, particularly for the highly noisy input at the initial step, poses a significant challenge and incurs additional training costs.

% Even though the classifier guidance has caused a performance improvement to some extent. However, it requires training an extra classifier, and it is not easy to train robust classifiers for all reverse steps with varying levels of noisy images, especially for the highly noisy input at the initial step. 
% As a result, such a classifier can potentially lead the generation process astray.
%from  $p_{\theta}(x_{t-1}|x_{t}, c)$ 
%e it requires training an extra classifier, and this classifier must be trained on noisy data so it is generally not possible to plug in a pre-trained classifier

% In this case, the score model $\epsilon_{\theta}$ takes $c$ as another input to model the corresponding conditional score, i.e., $\epsilon_{\theta}(x_t,c,t)\approx -\sigma_t\nabla_{x_t}\log p(x_t|c)$.

To avoid training a separate classifier model, classifier-free guidance~\cite{ho2021classifier} takes $c$ as another input of the denoising neural network to model the conditional diffusion score, i.e., $\epsilon_{\theta}(x_t,c,t)\approx -\sigma_t\nabla_{x_t}\log p(x_t|c)$, while the unconditional score $\epsilon_{\theta}(x_t,t)$ is jointly estimated by randomly dropping the text prompt with a certain probability at each training iteration.
% classifier-free guidance~\cite{} propose to train a single neural network to model both unconditional and conditional diffusion score, i.e., $\epsilon_{\theta}(x_t,t)$ and $\epsilon_{\theta}(x_t,c,t)$, where the text prompt for each image is dropped randomly with a certain probability at each training iteration. 
Then the gradients for the classifier $p_{\phi}(c|x_t)$ can be estimated as{:}
\begin{equation}
    \begin{split}
        \nabla_{x_t} \log p(c|x_t) &= \nabla_{x_t} \log p_{\theta}(x_t|y) - \nabla_{x_t} \log p_{\theta}(x_t)\\
        &=-\frac{1}{\sigma_t} (\epsilon_{\theta}(x_t,c,t) - \epsilon_{\theta}(x_t, t)).
    \end{split}\label{equ:gradclass}
\end{equation}
Along this line, the corresponding diffusion score in Equation~\ref{equ:cg} can be derived as:
\begin{equation}
    \begin{split}
        \hat{\epsilon}_{\theta}(x_t. c, t) = \epsilon_{\theta}(x_t, t) + \gamma (\epsilon_{\theta}(x_t,c, t) -\epsilon_{\theta}(x_t, t) ),
    \end{split}\label{equ:cfg}
\end{equation}
where $\gamma$ is also usually set as a global scalar parameter to control the guidance degree of the condition. 
However, in this paper, we argue that the CFG scale should be spatially adaptive, allowing for balancing the inconsistency of semantic strengths for diverse semantic units in the image.

% we argue that the CFG scale should be spatial-ware to adaptively strengthen different semantic units consisting of the text prompt $c$.

% However, we argue that 
% each text prompt consists of semantic units, i.e., $c={c_1, ..., c_n}$. where each unit $c_i$ corresponds to relatively independent semantic information, such as a word token in the text prompt or its hyponymy and hypernymy. Take the 

%可视化 不同语义单元的norm delta norm变化 
% Suppose that $p_{\theta}{x_0}$ is the learned unconditional diffusion model.
% During inference, we wish to condition the generation on a label to text prompt $c$ 

%% file: sec/4_methods.tex
\begin{figure*}[t]
    \centering
    \includegraphics[width=0.95\linewidth]{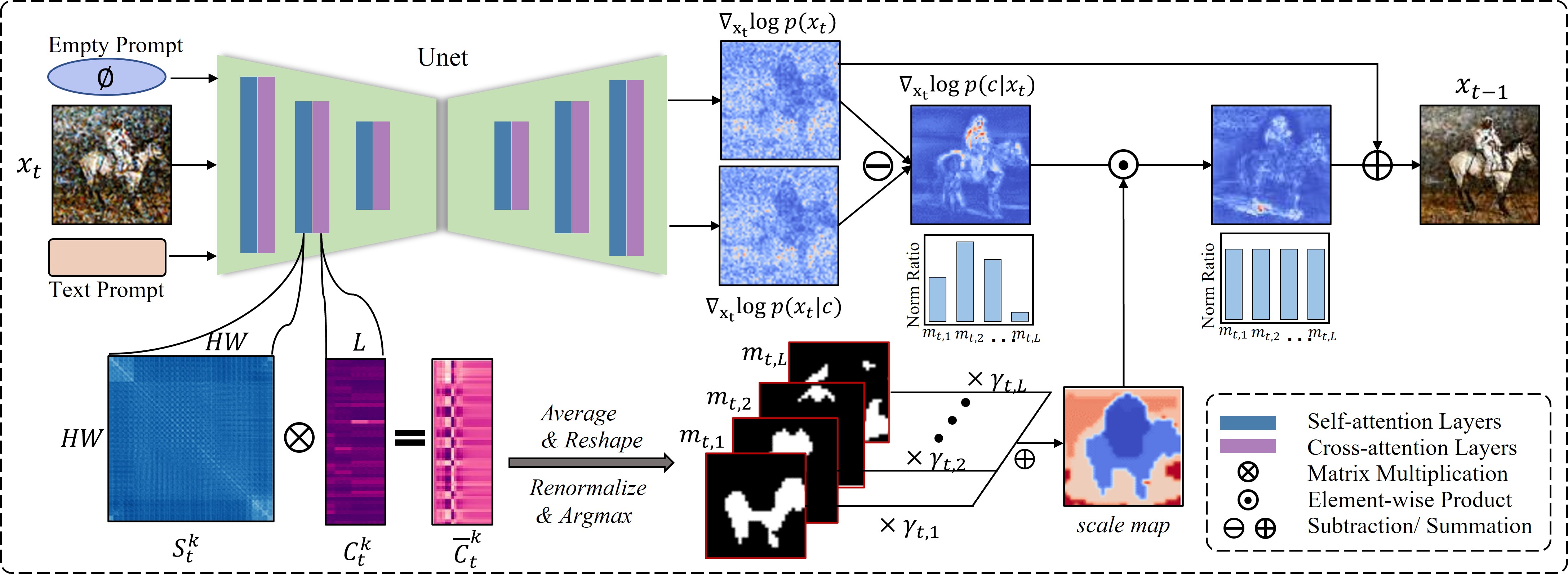}
    \vspace{-2mm}
    \caption{\textbf{The overall framework of our S-CFG method.} 
    At each denoising step in diffusion models, the U-net backbone estimates both diffusion score $\nabla_{x_t} \log p(x_t)$ and conditional diffusion score $\nabla_{x_t} \log p(x_t|c)$ without or with text prompt input, which can further infer the classifier score $\nabla_{x_t} \log p(c| x_t)$.
    By extracting and exploiting self-attention map $S^k_t$ and cross-attention map $C^k_t$ in each attention layer of U-net, we can obtain the region masks $m_{t,i}$ for each prompt token $i$. With the goal of unifying the classifier score norm in different regions, the CFG scale map can be determined to control the semantic strengths spatially in the following step.
}
    \label{fig:framework}
    \vspace{-4mm}
\end{figure*}

\vspace{-1mm}
\section{Methods}\label{sec:method}
\vspace{-2mm}
In this section, we introduce the technical details of Semantic-aware Classifier-Free Guidance (S-CFG). where the overview of the framework is shown in Figure~\ref{fig:framework}. 
At each denoising step in diffusion models, the current latent image is fed into the U-net backbone to estimate both diffusion score and conditional diffusion score without or with text prompt input.
With the extracted attention maps, we can derive region masks for the relatively independent semantic units.
In particular, the cross-attention map is renormalized for assigning each patch to the corresponding token, while the self-attention map is used to complete the semantic regions.
Then, to balance the amplification of diverse semantic information, we set adaptive CFG scales on diverse region masks and obtain the scale map to rescale their classifier scores into a uniform level.

\vspace{-1mm}
\subsection{Segmantic Map Generation}\label{sec:segmentation}
\vspace{-2mm}
To customizedly control the amplification of diverse semantic units, we need to segment the latent image once using the CFG strategy defined in Equation~\ref{equ:cfg}, i.e., at each denoising step.
However, this task is not trivial because the final image can not be accessed during the generation process. 
Fortunately, the attention layers in the U-net backbone have been reported to contain valuable semantic information for capturing relationships between image and text prompts~\cite{chefer2023attend,wang2023diffusion}, which can be leveraged to efficiently extract semantic units.

Specifically, for most text-to-image diffusion models, the interaction between the text prompt and the generation image is performed using cross-attention mechanisms. In general, the denoising U-net network consists of self-attention layers followed by cross-attention layers at certain resolutions. For example, SD puts 16 self- and cross-attention layers at the resolution of  64, 32, 16, 8. In {the $k$-th} attention layer, a self-attention map $S^k_t\in \mathbb{R}^{HW \times HW}$ and a cross-attention map $C^k_t\in \mathbb{R}^{HW \times L}$ are calculated over linear projections of the intermediate image spatial feature $z^k_t\in \mathbb{R}^{H W \times C}$ or text embedding $e\in\mathbb{R}^{L \times D}$, 
\begin{equation}
    \begin{split}
    S_t^k = {\rm Softmax} \left (\frac{Q_s(z^k_t)K_s(z^k_t)^T}{\sqrt{d}}\right ),\\
        C_t^k = {\rm Softmax} \left (\frac{Q_c(z^k_t)K_c(e)^T}{\sqrt{d}} \right ),
    \end{split}
\end{equation}
where $H$ and $W$ are the current resolutions, $L$ is the number of text tokens, $C$ is the image feature channel, $D$ is the token embedding  dimension, and $Q_{*}(\cdot)$ and $K_{*}(\cdot)$ are linear {projections} with the dimension of output as $d$. 
% Similarly, in $k$-th self-attention layer, an attention map $S_t^k\in \mathbb{R}^{HW\times HW}$ is derived by a similar equation to the above where $K$ and $Q$ are all based on the intermediate image spatial feature.

 \vspace{-1mm}
\subsubsection{Cross-Attention-based Semantic Segmentation}
\vspace{-2mm}
Intuitively, at each denoising step $t$, each row in $C_t^k$ defines the distribution over the text tokens, which is used to augment with the most relevant textual token for each {patch}. Therefore, a higher probability $C_t^k[s, i]$ indicates a closer relationship between the current patch $s$ and the corresponding token $w_i$.
Along this line, we propose to segment the latent image $x_t$ as the set of regions masked by $\{ m_{t,1}, ..., m_{t,L}\}$, where $i$-th masked region $m_{t,i}\in \{0,1\}^{HW}$ corresponds to the semantic token $w_i$.
% all text tokens-related regions  $c_{x,t}= \{c^{w}_{t,1},...,  c^{w}_{t, L}\}$, where each semantic unit $c^w_{t, i}$ corresponds to the $i$-th token $w_i$ in the text prompt $c$ and spatial region at this step $t$ with the mask $m^c_{t,i}\in \{0,1\}^{HW}$.
%i.e., $c^w_{t,i} = (w_i, m^w_{t,i})$. 

Specifically, we first employ a fusion process to obtain the final cross-attention map $C_t \in \mathbb{R}^{HW\times L}$. This fusion involves averaging the cross-attention layers and heads with the smallest two resolutions, as these have been shown to contain the most substantial semantic information~\cite{hertz2022prompt}. In particular, all attention maps are upsampled into the same size. 
Then, $C_t$ is renormalized along the spatial dimension, and the argmax operation is applied on the token dimension to determine the activation of the current patch, denoted as:
% to extract the spatial region activated by each token $w_i$ at step $t$
\begin{equation}
    \begin{split}
        \hat{C}_t[s, i] = \frac{C_t[s,i]}{\sum_{s'=1}^{HW} C_{t}[s',i]},\\
        i_s = \arg \max_i \hat{C}_t[s,i],
    \end{split}\label{equ:camap}
\end{equation}
where $\hat{C}_t[s, i]$ estimates the possibility assigned to the patch $s$ for the token $w_i$. The corresponding region mask $m_{t, i}$ can be derived by setting the element in the patch set $\{s: i_s=i\}$ as 1, and 0 for others.
Note that the renormalization in the above equation plays a crucial role in aligning the token with the image patch in our practice. Without the renormalization,  $C_t$ would tend to concentrate most of the attention on a single token, such as the START token, for all patches, damaging the semantic segmentation.

\begin{figure}
  \centering
  \includegraphics[width=0.98\linewidth]{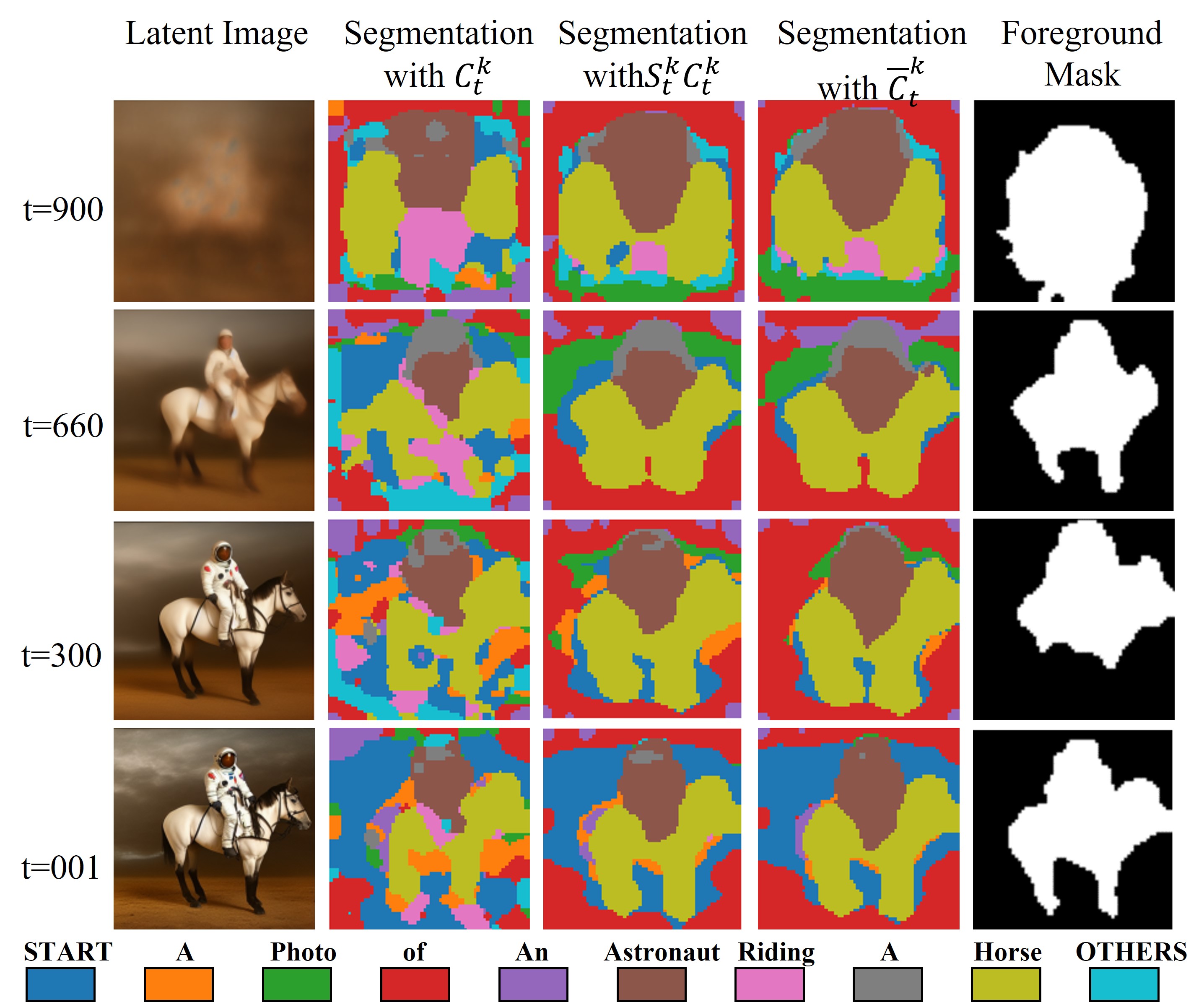}
  \vspace{-2mm}
  \caption{\textbf{The latent image segmentation based on attention maps at different denoising steps.} The first column shows the predicted image $x_0$ based on the current latent image $x_t$ and noise estimation $\epsilon_{\theta}$ with Equation~\ref{equ:xtx0}. The following three columns show the semantic segmentation maps with different strategies. Regions labeled by different colors correspond to different tokens. The last column shows the foreground mask detected by our approach. }
  \label{fig:seg}
  \vspace{-1mm}
\end{figure}

The second column in Figure~\ref{fig:seg} shows an example result of the above semantic segmentation, we can find that the semantic maps could successfully detect the rough locations of several important tokens, such as {``astronaut''} and ``horse''.
However, it is worth noting that they often exhibit unclear object boundaries and may contain internal holes, particularly during the initial denoising steps. To alleviate this problem, we propose to refine and complete the semantic map with self-attention maps in the following section.

\vspace{-1mm}
\subsubsection{Self-Attention-based Segmentation Completion}
\vspace{-2mm}
% The second column in Figure~\ref{fig:seg} shows an example of the above semantic segmentation, we can find that the semantic maps could successfully detect the rough locations of several important tokens, such as ``astronaut'' and ``horse''.
% However, it is worth noting that the segmentation maps often exhibit unclear object boundaries and may contain internal holes, particularly during the initial denoising steps.
% To alleviate this problem, 
Specifically, we follow~\cite{wang2023diffusion} and refine each cross-attention map $C_t^k$ by multiplying it with the corresponding self-attention maps at each attention layer. 
The hidden logic is rooted in the ability of self-attention maps to estimate the correlation between patches, enabling cross-attention to compensate for incomplete activation regions and perform region completion.
Meanwhile, note that $S_t^k$  can be interpreted as a transition matrix among all patches, where each element is nonnegative and the sum of each row equals 1. We can also enhance the region completion by {transmitting} semantic information among patches following the idea of feature propagation in graph~\cite{kipf2016semi}. Therefore, same as ~\cite{zhu2020simple}, we refine the cross-attention map $C_t^k$ as follows:
% $S^k_t$ as follows:
\begin{equation}
    \begin{split}
        % \overline{C}_t^k = \overline{S}_t^kC_t^k.\\
        % \overline{S}_t^k = \frac{1}{R}\sum_{i=1}^R (S_t^k)^i,\\
        \overline{C}_t^k = \frac{1}{R}\sum_{{r}=1}^R (S_t^k)^{r} C_t^k,
    \end{split}\label{equ:overlineC}
\end{equation}
% In particular, note that $S_t^k$  can be interpreted as a transition matrix among all patches, where each element is nonnegative and the sum of each row is equal to 1. We can also enhance the region completion by transiting semantic information among patches following the idea of feature propagation in graph~\cite{kipf2016semi}. In particular, same as ~\cite{zhu2020simple}, for each attention layer, we refine the $S^k_t$ as follows:
% \begin{equation}
%     \begin{split}
%         \overline{S}_t^k = \frac{1}{R}\sum_{i=1}^R (S_t^k)^i,
%     \end{split}\label{equ:overlineS}
% \end{equation}
where $R$ is a hyper-parameter and set as 4 in our {experiments}. 
Combining Eqaution~\ref{equ:overlineC}, a refined version of cross-attention map, i..e, $\overline{C}_t$, would be computed, which would be put into Equation~\ref{equ:camap} for deriving refined segmentation masks. The {fourth} column in Figure~\ref{fig:seg} shows the corresponding results, where segmentation maps become better with clearer object boundaries and fewer internal holes, even better than the third column which sets $R=1${.}

\vspace{-1mm}
\subsection{Semantic-Aware CFG }
\vspace{-2mm}
At each denoising step $t$, given the semantic units with masks $\{m_{t,1}, ..., m_{t,M}\}$, we turn to design the semantic-aware CFG strategy to control the strength of each semantic unit separately.
In particular, note that the image patches in the different semantic units usually have a more distant relationship than that among the same semantic unit.
To simplify the discussion, we assume that \textit{different semantic units are independent of each other at any time step}. Based on this assumption, we can derive the following { expressions} about the classifier $p(c|x_t)$:
\begin{equation}
    \begin{split}
        p(c|x_t) = \prod_{i=1}^{L} p(w_{i}|m_{t,i}\odot x_t),\\
        \nabla_{x_t} \log p(w_{i}|m_{t,i}\odot x_t) = m_{t,i} \odot \nabla_{x_t} \log p(c| x_t),
    \end{split}
\end{equation}
where $m_{t,i}$ is interpolated and reshaped to the same size as $x_t$ and $\odot$ is the element-wise product. (The detailed derivation can be found in the Appendix.)
Then, instead of using a single scalar to control the guidance degrees of all semantic units, like that in Equation ~\ref{equ:cg} and~\ref{equ:cfg}, we define the composed diffusion score function as follows:
\begin{equation}
    \begin{split}
        \hat{\epsilon}_{\theta}(x_t,& c,t) =\epsilon_{\theta}(x_t, t)  \\
        &+  \sum_{i=1}^M \gamma_{t,i} m_{t,i}\odot (\epsilon_{\theta}(x_t,c, t) -\epsilon_{\theta}(x_t, t)),
    \end{split}\label{equ:spatialcfg}
\end{equation}
where each term in the sum operation is the estimation of log-density for each semantic token $w_{i}$, and $\gamma_{t,i}$ is the scalar parameter to strengthen the corresponding semantic information. 
In particular, when all parameter $\gamma_{t,i}$ is set as the same as $\gamma$, the above equation {reduces into} the same as the original CFG strategy in Equation~\ref{equ:cfg}.

\begin{figure*}[t]
  \centering
  % \subfigure[]{}
  \begin{subfigure}{0.3\linewidth}
  \includegraphics[width=1.0\linewidth]{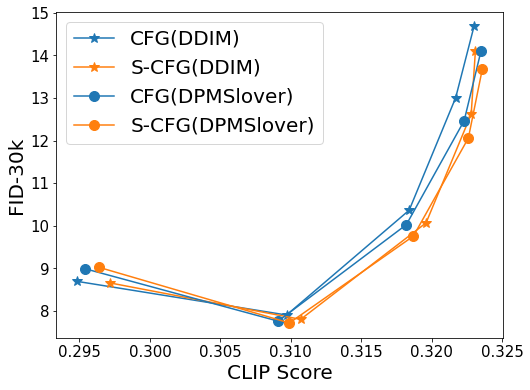}
  \caption{SD-v1.5}
  \end{subfigure} \hspace{3mm}
  \begin{subfigure}{0.285\linewidth}
  \includegraphics[width=1.0\linewidth]{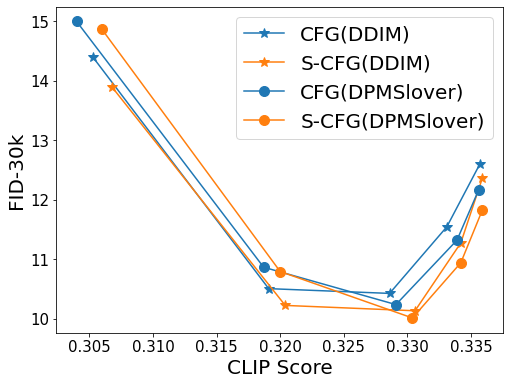}
  \caption{SD-v2.1}
  \end{subfigure} \hspace{3mm}
  \begin{subfigure}{0.285\linewidth}
  \includegraphics[width=1.0\linewidth]{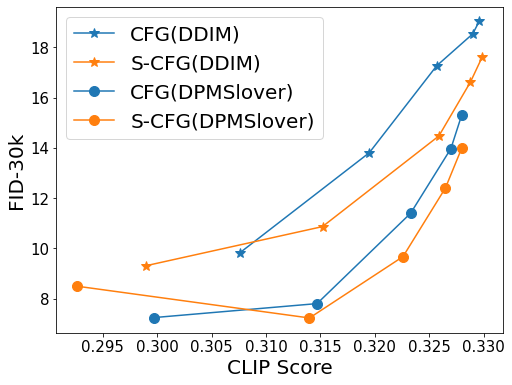}
  \caption{DeepFloyd IF}
  \end{subfigure}\\
  \vspace{-2mm}
  \caption{\textbf{The qualitative evaluation results on the trade-off curve of FID-30K VS CLIP Score.}}
  \vspace{-2mm}
  \label{fig:result-qual}
\end{figure*}

\vspace{-1mm}
\subsubsection{Adaptive CFG Scale $\gamma_{t,i}$}
\vspace{-2mm}
Here, we further propose an approach to adaptively set the CFG scale $\gamma_{t,i}$. 
The primary objective is to achieve a balanced amplification of diverse semantic units during each denoising step.  
To achieve this, an intuitive idea is to rescale the classifier scores in different semantic regions to a benchmark scale. This ensures that all semantic units undergo a comparable magnitude of change throughout the denoising process. Specifically, $\gamma_{t,i}$ is defined as follows:
% To achieve this, an intuitive idea is to rescale the gradient of classifiers in different semantic units to the same norm, which constrains that all set mantic units have the same magnitude of change during each denoising step. Specifically, we formulate $\gamma_{t,i}$ as follows:
\begin{equation}
    \begin{split}
        \eta_t &=\Vert \epsilon_{\theta}(x_t,c, t) -\epsilon_{\theta}(x_t, t) \Vert_2 \in \mathbb{R}^{HW}, \\
        \gamma_{t,i} &= \gamma \frac{|m_{t,b} \odot \eta_t|}{|m_{t,i}\odot \eta_t|}\frac{|m_{t,i}|}{|m_{t,b}|},
    \end{split}\label{equ:spatialgamma}
\end{equation}
where $\Vert \cdot \Vert_2$ is the 2-norm operator of vectors used on the last dimension of a tensor, and $|\cdot |$ is the sum operator of a vector or matrix. $\gamma$ is a hyper-parameter shared for all samples and time steps, like that in the original CFG strategy.  In particular, the mask $m_{t,b}\in \{0,1\}^{HW}$ is introduced to assign the benchmarking region. For example, when setting $m_{t,b}$ as 1 for any patch, the average patch norm of the current latent image is the benchmark scale. Here we also introduce another benchmark region for better performance, i.e., the foreground region, such as the union of the regions of  ``astronaut'' and ``horse'' in Figure~\ref{fig:example}.

Specifically, when estimating the unconditional score $\nabla_{x_t}\log p(x_t)$, an empty prompt $\emptyset$ is {fed} into the model, i.e,  $\epsilon_{\theta}(x_t,\emptyset, t)$, where $\emptyset$ is usually represented as a list of padding tokens with a start token. Based on our approach in Section~\ref{sec:segmentation}, we can detect the semantic region of the START token $m_{t, \text{START}}$, which effectively indicates the background area in our implementation (see the last column in  Figure~\ref{fig:seg}).
Therefore, { we can align the benchmarking region with the foreground region by setting}:
\begin{equation}
    \begin{split}
        m_{t,b} = 1-m_{t, \text{START}}.
    \end{split}\label{equ:fore}
\end{equation}

%% file: sec/5_experiment.tex
\vspace{-2mm}
\section{Experiments}
\vspace{-1mm}
% To verify the effectiveness of the proposed method, S-CFG, we systematically conduct a series of experiments, applying our method to state-of-the-art text-to-image diffusion models, with quantitative, human-level, and qualitative evaluations. 
% Albation analysis and downstream tasks are also discussed.
% Notably, our approach operates in a training-free manner without the need for supplementary training procedures.
% To assess our method, we apply S-CFG to state-of-the-art text-to-image diffusion models and evaluate its performance through quantitative, human-level, and qualitative assessments. Additionally, we conduct ablation analysis and evaluate the impact of our method on downstream tasks.
% Importantly, our approach operates in a training-free manner, eliminating the need for additional training procedures.

\noindent\textbf{Benchmark Models.} 
 We include two diffusion models as base models: {Stable diffusion (SD)~\cite{rombach2022high}}, which operates in the latent image space, and {DeepFloyd IF (IF)~\cite{Shonenkov2023Deep}}, which operates in the image pixel space. 
Specifically, we consider two versions of SD: SD-v1.5 and SD-v2.1, which differ in terms of model sizes and generative qualities. For the IF model, we use the middle-scale version, IF-M, which is constructed using multiple diffusion models.
To maintain simplicity, two model stages are used, where the base diffusion model produces low-resolution samples and an upscale diffusion model boosts them to a higher resolution. Both stages can benefit from the CFG or S-CFG strategy.
Additionally, the IF model uses the T5XXL as the text encoder without using the start token. Therefore, instead of assigning the foreground region based on the start token, we set the benchmarking mask $m_{t,b}$ in Equation~\ref{equ:spatialgamma} as 1 for any patch.
All three models are publicly accessible.

Meanwhile, two samplers are discussed for all three models, i.e., DDIM~\cite{wang2023diffusion} and {DPMSolver}++~\cite{lu2022dpm2}, which are both the most widely used in practice.
Specifically, for DDIM, we follow~\cite{rombach2022high} and set the number of sampling steps as 250 for SD models with the noise variance parameter as 0.
Regarding the IF model, which employs learnable noise variance parameters, we adhere to the original noise settings and conduct DDIM sampling with 50 steps.
As for {DPMSolver}++, we set the number of sampling steps as 50.

\begin{figure*}[t]
  \centering
  % \subfigure[]{}
  \includegraphics[width=0.95\linewidth]{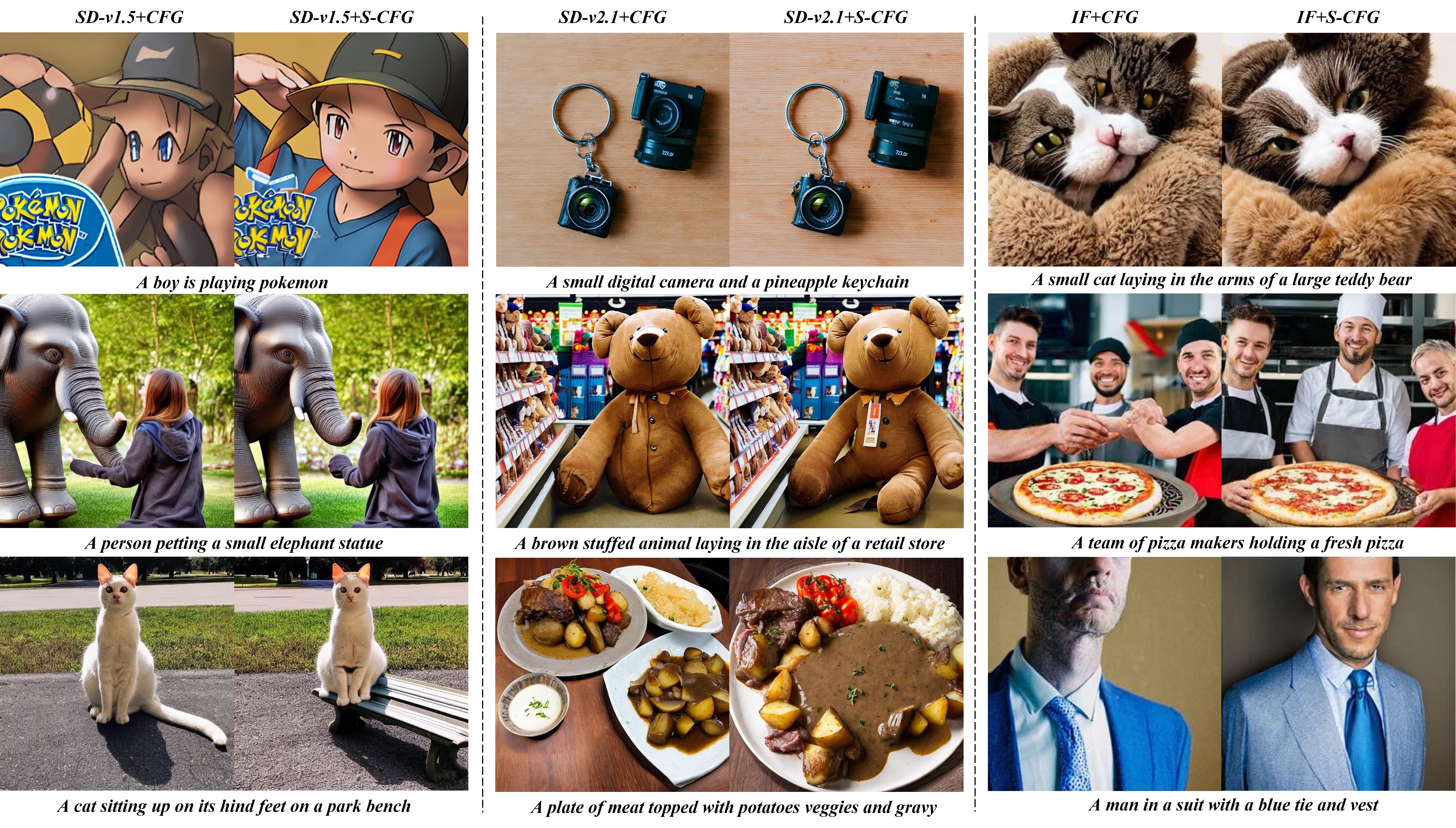} %\hspace{2mm}
  \vspace{-2mm}
  \caption{\textbf{Samples generated by different base models with CFG (left) or S-CFG (right).}}
  \vspace{-4mm}
  \label{fig:case}
\end{figure*}

% \begin{figure*}[t]
%   \centering
%   % \subfigure[]{}
%   \includegraphics[width=0.31\linewidth]{figs/SD15-case-1.png} %\hspace{2mm}
%   \includegraphics[width=0.31\linewidth]{figs/SD21-case-3.png}
%   \includegraphics[width=0.31\linewidth]{figs/IF-case-1.png}
%  \\
%  \includegraphics[width=0.31\linewidth]{figs/SD15-case-2.png}
%   \includegraphics[width=0.31\linewidth]{figs/SD21-case-2.png}
%   \includegraphics[width=0.31\linewidth]{figs/IF-case-2.png}
%   \\
%   \begin{subfigure}{0.31\linewidth}
%   \includegraphics[width=1.0\linewidth]{figs/SD15-case-5.png}
%   \caption{SD-v1.5}
%   \vspace{-2mm}
%   \end{subfigure}
%   \begin{subfigure}{0.31\linewidth}
%   \includegraphics[width=1.0\linewidth]{figs/SD21-case-4.png}
%   \caption{SD-v2.1}
%   \vspace{-2mm}
%   \end{subfigure}
%   \begin{subfigure}{0.31\linewidth}
%   \includegraphics[width=1.0\linewidth]{figs/IF-case-3.png}
%   \caption{DeepFloyd IF}
%   \vspace{-2mm}
%   \end{subfigure}
%   \caption{\textbf{Samples generated by different base models with CFG (left) or S-CFG (right).}}
%   \vspace{-4mm}
%   \label{fig:case}
% \end{figure*}

\vspace{-1mm}
\subsection{Quantitative  Evaluation}
\vspace{-2mm}
We compare the benchmark models with CFG and S-CFG on the MSCOCO 256$\times$ 256 dataset.  Two qualitative metrics are used: 1) FID-30K: zero-shot Frechet Inception Distance with 30K images and the corresponding captions, which measures the quality and diversity of images. 2) CLIP Score~\cite{radford2021learning}:  which randomly selects 5K captions as prompts and uses the CLIP model to assess the alignments between the generated images and their corresponding text prompts. 
In particular, the trade-off between FID and CLIP scores has been widely reported with varying CFG scales~\cite{nichol2021glide}. Therefore, we present the trade-off curve across a range of the global scale $\gamma \in [2.0,~3.0,~5.0,~7.5,~10.0]$.

% Figure~\ref{fig:result-qual} shows the results 
Based on the results presented in Figure~\ref{fig:result-qual}, it is evident that our S-CFG strategy consistently outperforms the original CFG strategy across most experimental settings, where the trade-off curve of S-CFG  consistently favors a position towards the bottom right of that of the original CFG strategy in each setting (See Appendix for a full detailed table).
This phenomenon demonstrates the effectiveness and robustness of S-CFG, establishing its applicability in both latent image space and pixel space for diffusion models with different model sizes.
In addition, we can find that the diffusion sampler may be crucial for the generative quality, specifically for the pixel space model, i.e., IF, where a significant performance gap is observed for DDIM and  {DPMSolver}++.  However, S-CFG also achieve performance improvement.
% In particular, it is somewhat weird that SD-v2.1 fails to outperform SD-v1.5 in our settings. We also reproduce the performance comparisons with a similar setting as the official report with FID-10K metric and DDIM with 50 steps. Our S-CFG method also achieves the best (See Appendix for more details).

% IF  1.5的情况

% Long and short captios 
\vspace{-1mm}
\subsection{Human-Level Evaluation}
\vspace{-2mm}
Here, 80 prompts are randomly selected from MSCOCO validation dataset for generative images with CFG and S-CFG. Then, we asked 5 participants to assess both the image quality and image-text alignment. 
%we employ DrawBench~\cite{saharia2022photorealistic},  a comprehensive and challenging benchmark for testing different capabilities of text-to-image models. 
Human raters are asked to select the superior respectively from the given two synthesized images, one from the original CFG strategy, and another from our S-CFG strategy.
For fairness, we use the same random seed for generating both images.
The voting results are summarised in Table~\ref{tab:human}. The majority of votes go to our S-CFG strategy for all base models, demonstrating superiority in both evaluated aspects.

\begin{table}[t]
\caption{\textbf{Human-level evaluation results.}}\label{tab:human}
\vspace{-2mm}
\begin{tabular}{lcccc}
\hline\hline
        & \multicolumn{2}{c}{Image Quality} & \multicolumn{2}{c}{Image-Text} \\\hline
        & CFG             & S-CFG           & CFG            & S-CFG         \\\hline
SD-v1.5 & 26.78\%          & \textbf{73.22}\%          & 23.20\%         & \textbf{76.80}\%        \\
SD-v2.1 & 28.16\%          & \textbf{71.84} \%          & 31.85\%         & \textbf{68.15}\%        \\
IF      & 32.39\%          & \textbf{67.61}\%          & 29.17\%         & \textbf{70.83}\%     \\
\hline\hline
\end{tabular}
\end{table}

\vspace{-1mm}
\subsection{Qualitative Evaluation}
\vspace{-2mm}
In Figure~\ref{fig:case}, we show some samples generated by different models with CFG and S-CFG. For fairness, we use the same setting and random seed for different strategies.
The results exhibit a notable enhancement in the model’s generative capacity from the aspects of semantic expressiveness and entity portrayal.
For example, when given the prompt ``A boy is playing Pokemon'', S-CFG improves SD-v1.5 by ensuring the boy's appearance in a normal manner.  In the case of ``A person petting a small elephant statue'', S-CFG eliminates the irregular elephant's trunk.
Similar improvement in fine-grained structure completion can also be observed for {SD-v2.1} and IF in the first two rows.
Furthermore, for scenarios in the last rows, such as  ``A cat sitting ... on a park bench'', ``A plate of meat topped ...'' and ``A man in a suit with a blue tie ...'', S-CFG helps models generate images that accurately represent the semantic descriptions.

\begin{figure}[t]
  \centering
  \includegraphics[width=0.6\linewidth]{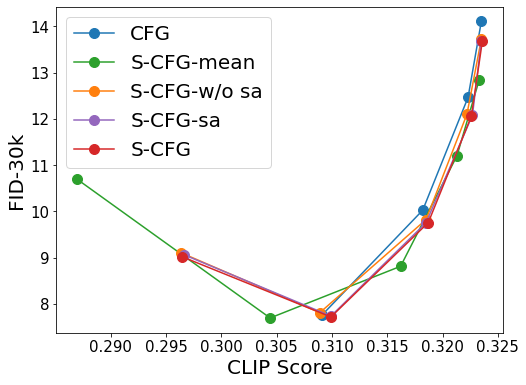}
  \caption{\textbf{The ablation analysis by evaluating the performance of different components in S-CFG.}}
  \label{fig:ablation}
  \vspace{-2mm}
\end{figure}
\vspace{-2mm}
\subsection{Ablation Analysis}
\vspace{-1mm}
Here, three variants of S-CFG are introduced: 1) S-CFG-mean sets the benchmarking mask $m_{t,b}$ as 1 for all patches. 2) S-CFG w/o sa is the variant without the segmentation completion based on self-attention maps. 3) S-CFG-sa is the variant with $R=1$ in Equation~\ref{equ:overlineC}.

The results in Figure~\ref{fig:ablation} based on SD-v1.5 demonstrate that all variants of S-CFG consistently outperform the original CFG strategy.
% particularly when considering large CLIP Scores (corresponding to larger $\gamma$ values). 
This observation strongly supports our core idea of setting customized CFG scales for different semantic regions throughout the denoising process.
In addition, when compared to other variants, S-CFG-mean exhibits increased performance instability and fails to achieve the optimal CLIP Score at the lowest FID score. It verifies the advantage of using the foreground region described in Equation~\ref{equ:fore} as the benchmarking region.
Meanwhile, S-CFG w/o sa falls short in outperforming S-CFG-sa and S-CFG, albeit by a relatively small margin. This outcome highlights the effectiveness of self-attention-based segmentation completion.
Furthermore, while S-CFG-sa and S-CFG demonstrate similar performance levels, Figure~\ref{fig:seg} shows that S-CFG exhibits superior segmentation capability, which should result in more accurate image generation. However, these improvements may not be fully captured by the current evaluation metrics.

% Lastly, S-CFG-sa and S-CFG demonstrate comparable performance levels. However, as depicted in Figure~\ref{fig:seg}, S-CFG exhibits superior segmentation capability, leading to more accurate image generation. These improvements may not be adequately captured by the current evaluation metrics.

\begin{table}[t]
\caption{\textbf{Performance comparisons of ControlNet with CFG and S-CFG}, where the base model is SD-v1.5, the parameter $\gamma=3.0$ and that sampler is {DPMSolver}++ with 50 steps.}\label{tab:controlnet}
\vspace{-2mm}
\begin{tabular}{lcccc}
\hline\hline
           & \multicolumn{2}{c}{FID} & \multicolumn{2}{c}{CLIP Score} \\\cline{2-5}
           & CFG      & S-CFG    & CFG     & S-CFG         \\\hline
Canny        & 8.670     &\textbf{8.382}        & 0.3006    & \textbf{0.3019}           \\
Segmentation     &  9.595 & \textbf{9.549}  & 0.3004  & {\textbf{0.3017}}          \\
\hline\hline
\end{tabular}
\end{table}

\vspace{-2mm}
\subsection{Downstream tasks}
\vspace{-1mm}
Here, we extend the evaluations from foundational image generation to more specialized downstream tasks.

\begin{figure}[t]
  \centering
  % \subfigure[]{}
  \includegraphics[width=0.95\linewidth]{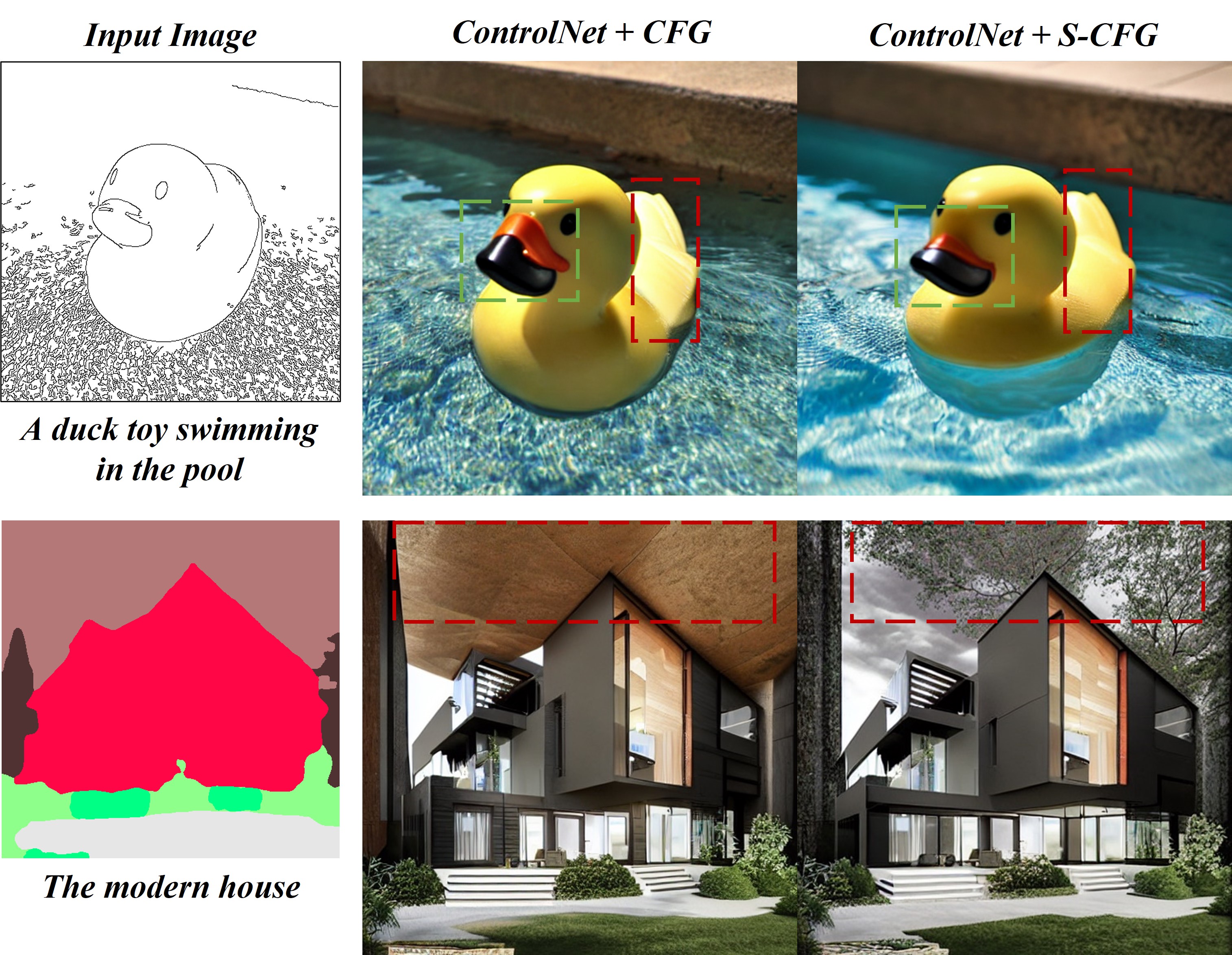}
  \vspace{-2mm}
  \caption{\textbf{Samples generated by {ControlNet} with CFG (middle) or S-CFG (right).} }
  \label{fig:controlnet}
  \vspace{-2mm}
\end{figure}

First, we incorporate S-CFG into ControlNet~\cite{zhang2023adding}, which is a neural network architecture for adding various spatial conditioning controls to text-to-image diffusion models. 
Specifically, we utilize SD-v1.5 as the base model, incorporating image canny edge and image segmentation as the spatial conditions. Table~\ref{tab:controlnet} presents a performance comparison between CFG and S-CFG. 
% Notably, we set the parameter $\gamma=3.0$ and employ {DPMSolver}++ with 50 steps as the sampler, as it performs the best in Figure~\ref{fig:result-qual}. 
The results demonstrate consistent improvement with the incorporation of S-CFG. 
Some examples are illustrated in Figure~\ref{fig:controlnet}, showcasing notable improvements in image realism. Specifically, in the canny case of the duck toy, S-CFG enhances the structure of the duck's mouth and rectifies color imbalances around the tail. %Moreover, the background pool exhibits more realistic water waves.
Likewise, in the segmentation case of the house, the ControlNet with CFG fails to synthesize the background sky, whereas S-CFG successfully addresses this issue.

% Some instances are shown in Figure~\ref{fig:controlnet}. Marked improvement can be witnessed in the image realism. Specifically, as for the canny case of the duck toy, S-CFG enhances the structure of the duck's mouth and eliminates the imbalance in color around the tail. Meanwhile, the pool in the background has also been changed with more normal water waves. 
% Similarly, for the segmentation case of the house, the original  ControlNet with CFG fails to synthesize the background sky, while S-CFG corrects that.

\begin{figure}[t]
  \centering
  % \subfigure[]{}
  \includegraphics[width=0.95\linewidth]{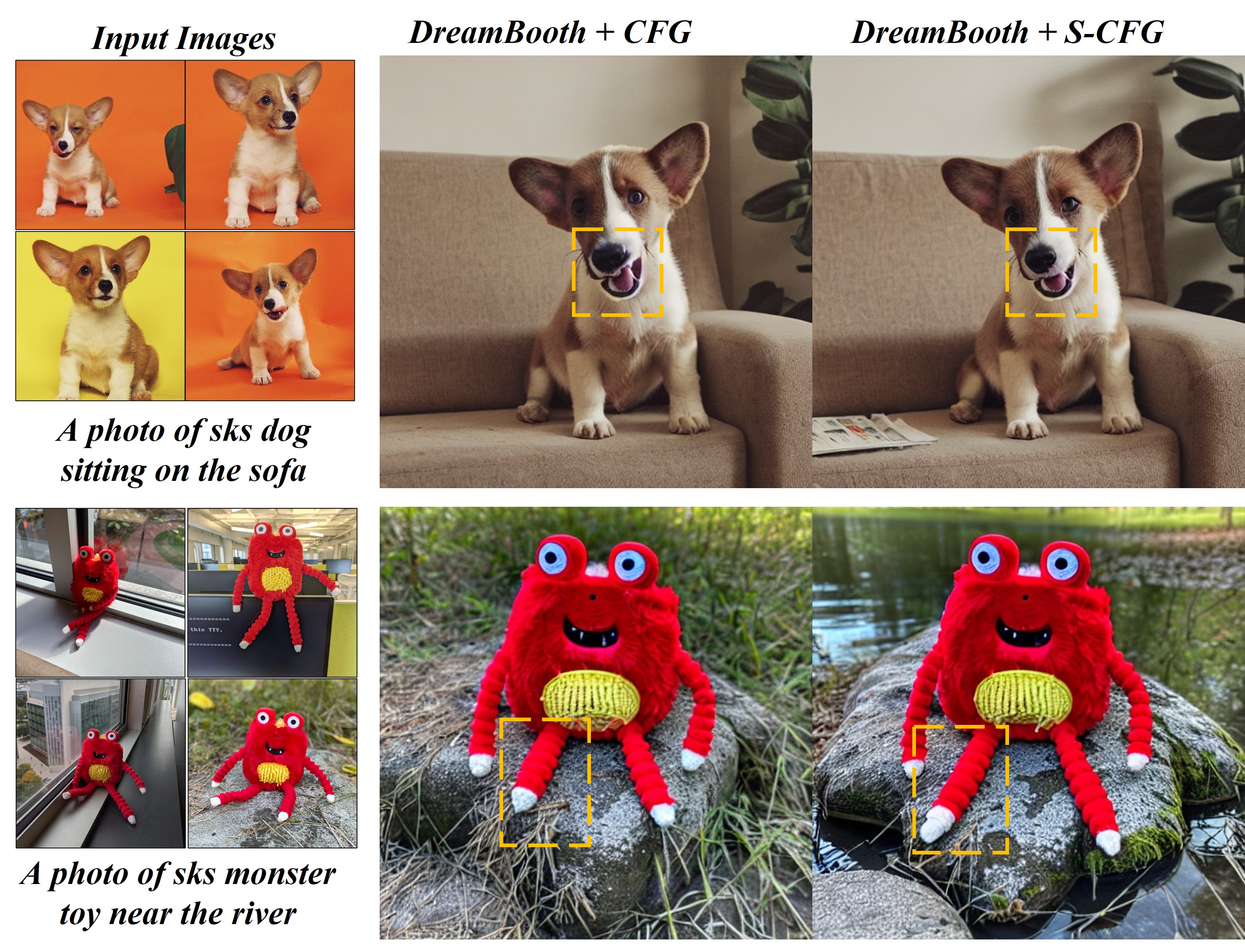}
  \vspace{-2mm}
  \caption{\textbf{Samples generated by DreamBooth with CFG (middle) or S-CFG (right).} {The token ``sks'' represents the shared subject among the input images.}}
  \label{fig:dreambooth}
  \vspace{-1mm}
\end{figure}

We have also integrated S-CFG into DreamBooth~\cite{ruiz2023dreambooth}, which enables the personalization of text-to-image diffusion models with specific subjects using only a few subject images. The examples presented in Figure~\ref{fig:dreambooth} highlight the improvements in image quality and text-image alignment achieved by S-CFG.
For instance, S-CFG enhances the appearance of the dog's mouth and brings the length of the toy's legs closer to the input images.
Notably, in the second row, DreamBooth with CFG fails to align the image with the text prompt {``river''}, 
whereas S-CFG succeeds.

%主试验

% 参数实验
%

%消融实验
% 不用 self-atten  [:] 只用一层 self-attn [-3：] 使用 mean norm 

%% file: sec/6_conclusion.tex
\vspace{-1mm}
\section{Conclusion}
\vspace{-2mm}
This paper argues that classifier-free guidance (CFG) in text-to-image diffusion models suffers from spatial inconsistency in semantic strengths and suboptimal image quality.
To this end, we proposed Semantic-aware CFG (S-CFG), customizing the guidance degrees for different semantic units.
Specifically, 
% by extracting the attention maps from the U-net backbone in diffusion models, 
we first design a training-free semantic segmentation method to partition the latent image into relatively independent semantic regions at each denoising step. 
% The cross-attention map in the denoising U-net backbone was renormalized for assigning each patch to the corresponding token, while the self-attention map was used to complete the semantic regions.
Then,  the CFG scales across regions are adaptively adjusted to rescale the classifier scores into a uniform level. 
% Finally, 
Experiments on multiple diffusion models demonstrated the superiority of S-CFG.

\vspace{-2mm}
\section{Acknowledgments}
\vspace{-2mm}
This research was supported by grants from the National Key R\&D Program of China (No. 2022ZD0119302).

%% file: sec/X_suppl.tex
\clearpage
\setcounter{page}{1}
\maketitlesupplementary

\section{Deriving Equation 11}
In this section, we provide a derivation for Equation 11 based on one assumption that may be not particularly strict, i.e., \textit{for any denoising step $t$, the semantic units, corresponding to token set $\{w_1,..., w_L\}$, with masks $\{ m_{t,1}, ..., m_{t,L}\}$ are independent of each other}. Along this line, we can derive:
\begin{equation}
    \begin{split}
     p(w_i|x_t) 
    &=\ p(w_i | \sum_{j=1}^L m_{t,j}\odot x_t)\\
    &=\frac{ \prod_{j=1}^L p( m_{t,j}\odot x_t |w_i) p(w_i)}{\prod_{j=1}^L p( m_{t,j}\odot x_t )} \\
    &= \frac{ p(m_{t,i}\odot x_t | w_i)p(w_i)\prod_{j=1, j\neq i}^L p( m_{t,j}\odot x_t) }{\prod_{j=1}^L p( m_{t,j}\odot x_t )} \\
    &=  \frac{ p(m_{t,i}\odot x_t | w_i)p(w_i) }{ p( m_{t,i}\odot x_t )}\\
    &=    p(w_i | m_{t,i}\odot x_t). \nonumber
    \end{split}
\end{equation}

Then, we can deduce Equation 11 as follows:
\begin{equation}
    \begin{split}
        p(c|x_t) &= \prod_{i=1}^L p(w_i|x_t) \\
        &= \prod_{i=1}^L p(w_i | m_{t,i}\odot x_t).\\
        \nabla_{x_t} \log p(w_i|m_{t,i}\odot x_t) &= \nabla_{m_{t,i}\odot x_t} \log p(w_i|m_{t,i}\odot x_t) \\
        &= \nabla_{m_{t,i}\odot x_t} \log p(w_i| x_t) \\
        & = \nabla_{m_{t,i}\odot x_t} \log p(c| x_t) \\
        &= m_{t,i} \odot \nabla_{x_t} \log p(c| x_t). \nonumber
    \end{split}
\end{equation}

% \begin{equation}
%     \begin{split}
%     p(c|x_t) = \prod_{i=1}^L p(w_i|x_t) 
%     =\prod_{i=1}^L p(w_i | \sum_{j=1}^L m_{t,j}\odot x_t)\\
%     =\prod_{i=1}^L \frac{ \prod_{j=1}^L p( m_{t,j}\odot x_t |w_i) p(w_i)}{\prod_{j=1}^L p( m_{t,j}\odot x_t )} \\
%     = \prod_{i=1}^L \frac{ p(m_{t,i}\odot x_t | w_i)p(w_i)\prod_{j=1, j\neq i}^L p( m_{t,j}\odot x_t) }{\prod_{j=1}^L p( m_{t,j}\odot x_t )} \\
%     = \prod_{i=1}^L \frac{ p(m_{t,i}\odot x_t | w_i)p(w_i) }{ p( m_{t,i}\odot x_t )}=  \prod_{i=1}^L  p(w_i | m_{t,i}\odot x_t)
%     \end{split}
% \end{equation}
Note that the prior assumption may not be strict in practice. However, it is intuitive that the patches among different semantic regions are more independent than those in the same patches.  Meanwhile, based on the segmentation examples in  Figure 3 and our experimental results, we believe that it is beneficial to segment the latent image and customize guidance degrees for different semantic regions.
% In addition, we are sorry for the typo in Equation 11, where the sign $M$ should be replaced by $L$ for the consistency of the notation.

\section{More Experimental Details}

\noindent\textbf{Benchmark Models.} In our experiment, we involve three special diffusion models as the benchmarks, which are all publicly accessible:
\begin{itemize}
    \item Stable Diffusion v1.5 (\textbf{SD-v1.5}), a diffusion model in the latent space of powerful pre-trained autoencoders~\footnote{https://huggingface.co/runwayml/stable-diffusion-v1-5}, which use the CLIP~\cite{radford2021learning} as the text encoder and output images with the resolution 512x512.  
    \item Stable Diffusion v2.1 (\textbf{SD-v2.1}), a variant of SD-v1.5 with more model size~\footnote{https://huggingface.co/stabilityai/stable-diffusion-2-1}, which can output images with the resolution 768$\times$768. 
    \item DeepFloyd IF (\textbf{IF}), is a diffusion model in the pixel image space~\footnote{https://huggingface.co/DeepFloyd/IF-I-M-v1.0}, which is constructed using multiple diffusion models with T5XXL as the text encoder. In particular, we use the first two stages of the middle-scale version, i.e., IF-I-M-v1.0 and IF-II-M-v1.0, which produce the 64$\times$64 resolution image and boost them into 256$\times$ 256 resolution, respectively. 
\end{itemize}

\noindent\textbf{Quantitative Metric.}
Two qualitative metrics based on the MSCOCO validation dataset are used:
\begin{itemize}
    \item \textbf{FID-30K}, where the FID score is computed on the 30K generated images with prompts selected from the validation set and the corresponding original images.
    \item \textbf{CLIP Score}, where 5K captions are selected randomly for guiding image synthesis, and CLIP-VIT-G-14~\footnote{https://huggingface.co/laion/CLIP-ViT-g-14-laion2B-s34B-b88K} is used to compute the similarity between the generated image and the corresponding caption.
\end{itemize}

\begin{figure*}[h]
    \centering
    \begin{subfigure}{0.285\linewidth}
  \includegraphics[width=1.0\linewidth]{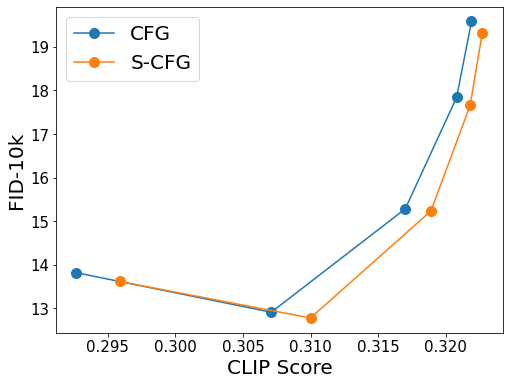}
  \caption{SD-v1.5}
  \end{subfigure}
  \begin{subfigure}{0.3\linewidth}
  \includegraphics[width=1.0\linewidth]{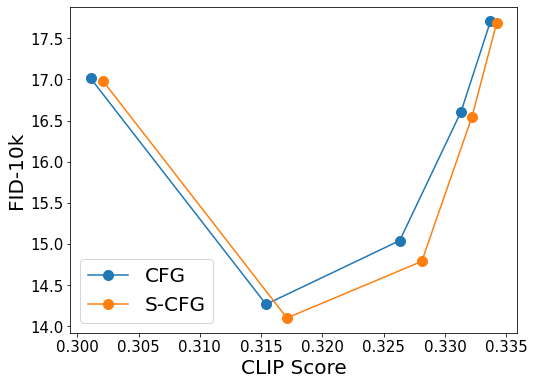}
  \caption{SD-v2.1}
  \end{subfigure}
  \vspace{-2mm}
    \caption{\textbf{The trade-off curve of FID-10K VS CLIP Score with DDIM sampler.}}
    \vspace{-2mm}
    \label{fig:fid10k}
\end{figure*}

In particular, our metric settings may be different from those in the official reports of the SD and IF models. It is somewhat weird that SD-v2.1 fails to outperform SD-v1.5 in our settings. Here, we also add another comparison on them based on a similar setting to their official report~\footnote{https://huggingface.co/stabilityai/stable-diffusion-2}, i.e., where FID-10k and CLIP Score (CLIP-VIT-G-14) on MSCOCO dataset are used with the 50-step DDIM sampler. The results are shown in Figure~\ref{fig:fid10k}. We can find that our S-CFG strategy also outperforms the original CFG strategy.

% In particular, we will publish our code for better reproduction after the possible acceptance. 

\section{Analysis on the Efficiency }
Here, we provide an additional analysis of the time cost of our S-CFG strategy. Specifically, we use { DPMSolver}++ with 50 steps as the sampler to generate images with different base models. All programs run on a single A100 GPU. Table~\ref{tab:timecost} shows the average time cost for generating a sample in 10 runs. We can find only a tiny time cost has been required compared with the original CFG strategy.

\begin{table}[h]
\centering
\caption{\textbf{The analysis on the time cost.}}\label{tab:timecost}
\begin{tabular}{lccc}\hline\hline
        & CFG   & S-CFG & improv. \\ \hline
SD-v1.5 & 2.773 & 2.848 & 2.70\% \\
SD-v2.1 & 7.054 & 7.167 & 1.60\% \\
IF      & 8.595 & 8.847 & 2.93\% \\
\hline\hline
\end{tabular}
\end{table}

\vspace{-2mm}
\section{More Ablation Analysis}
\begin{figure}
    \centering
    \includegraphics[width=0.6\linewidth]{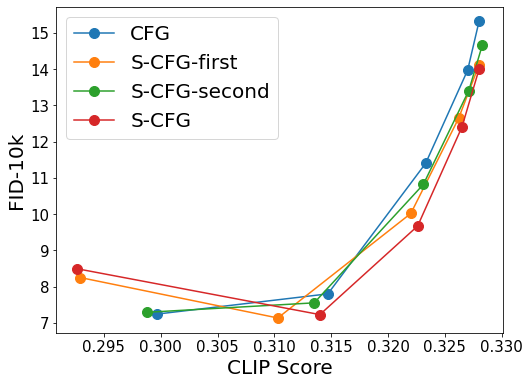}
    \caption{\textbf{The ablation analysis of the S-CFG on the diffusion model with multiple stages.}}
    \label{fig:ifablation}
\end{figure}

Here, we provide an additional ablation analysis of the S-CFG on the diffusion model with multiple stages, such as DeepFloyd IF~\cite{Shonenkov2023Deep}. We try to respond to the question: \textit{should the S-CFG strategy be used on all diffusion stages?} Specifically, based on the IF model used in our paper, we compare the performance of three methods:
\begin{itemize}
    \item \textbf{S-CFG-first}, where the S-CFG strategy is only used in the first diffusion model, i.e., IF-I-M-v1.0. 
    \item \textbf{S-CFG-second}, where the S-CFG strategy is only used in the second diffusion model, i.e., IF-II-M-v1.0.
    \item \textbf{S-CFG}, where the S-CFG strategy is used in both two diffusion models. 
\end{itemize}
In addition, the original CFG strategy is involved as a baseline. We use { DPMSolver}++ as the sampler with 50 steps and vary the parameter $\gamma$ in [2.0, 3.0, 5.0, 7.5, 10.0]. The trade-off curve of FID-30k VS CLIP Score is shown in Figure~\ref{fig:ifablation}. We can find that S-CFG tends to achieve the best trade-off between FID-30K and ClIP Score, while  S-CFG-first and S-CFG-second perform similarly.

\section{More Evaluation on Effectiveness}
 Recently,  a new metric called T2I-CompBench~\cite{huang2024t2i} was introduced to evaluate diffusion models, which assesses image quality from 6 aspects and aligns with human preference better. Here, we provide another comparison based on this metric. The results in Table~\ref{tab:exp} show that SD-v2.1 outperforms SD-v1.5 significantly, and S-CFG performs better than CFG.

\begin{table}[t]
\centering
\caption{ Evaluation on T2I-CompBench, where the $\gamma=7.5$.}\label{tab:exp}
\scalebox{0.70}{
\begin{tabular}{l|ccc|cc|c}
\hline\hline
\multicolumn{1}{c|}{\multirow{2}{*}{Model}} & \multicolumn{3}{|c|}{Attribute Binding}               & \multicolumn{2}{c|}{Object Relationship} & \multirow{2}{*}{Complex} \\
\multicolumn{1}{c|}{}                       & Shape           & Color           & Texture         & Non-Spatial        & Spatial            &                          \\\hline
SD-v1.5+CFG                                 & 0.3664          & 0.3761          & 0.4286          & 0.3109    & 0.111              & 0.2969                   \\
SD-v1.5+S-CFG                                   & \textbf{0.3793} & \textbf{0.3879} & \textbf{0.4288} & \textbf{0.3111}    & \textbf{0.1182}    & \textbf{0.2993}          \\\hline
SD-v2.1+CFG                                 & 0.4518          & 0.549           & 0.5146          & 0.3096             & 0.1512             & 0.3154                   \\
SD-v2.1+S-CFG                                 & \textbf{0.4558} & \textbf{0.5649} & \textbf{0.5333} & \textbf{0.3104}    & \textbf{0.1567}    & \textbf{0.3168} \\ %\hline
% KandinSky3+CFG  ($\gamma=5.0$)                                 & 0.5183          & 0.6325           & 0.5939         & 0.3145           & 0.2388             & 0.3317                   \\
% KandinSky+S-CFG ($\gamma=5.0$)                                 & \textbf{0.5215} & \textbf{0.6373} & \textbf{0.6005} & \textbf{0.3150}    & \textbf{0.2426}    & \textbf{0.3320} \\
\hline\hline
\end{tabular}}
\end{table}

\section{Detailed Table of Experiments}
Here, we show the detailed tables for experiments in Figures 4 and 6. We can find that our S-CFG achieves the best performance on all settings, with the best FID-30K score and CLIP Score.

\begin{table*}[h]
\caption{\textbf{The trade-off curve of SD-v1.5}, where the best FID-30k and CLIP Score are highlighted.}\label{tab:sdv15}
\begin{tabular}{c|cc|cc|cc|cc} \hline\hline
         & \multicolumn{4}{c}{DDIM}                                            & \multicolumn{4}{c}{{ DPMSolver}++}                                     \\\cline{2-9}
         & \multicolumn{2}{c}{CFG}          & \multicolumn{2}{c}{S-CFG}        & \multicolumn{2}{c}{CFG}          & \multicolumn{2}{c}{S-CFG}        \\\hline
$\gamma$ & FID-30K        & CLIP Score      & FID-30K        & CLIP Score      & FID-30K        & CLIP Score      & FID-30K        & CLIP Score      \\\hline
2.0      & 8.696          & 0.2948          & 8.656          & 0.2972          & 8.991          & 0.2954          & 9.023          & 0.2964          \\
3.0      & \textbf{7.904} & 0.3097          & \textbf{7.802} & 0.3107          & \textbf{7.760} & 0.3091          & \textbf{7.717} & 0.3099          \\
5.0      & 10.366         & 0.3184          & 10.069         & 0.3196          & 10.026         & 0.3182          & 9.757          & 0.3187          \\
7.5      & 13.008         & 0.3217          & 12.620         & 0.3228          & 12.466         & 0.3223          & 12.059         & 0.3226          \\
10.0     & 14.682         & \textbf{0.3230} & 14.101         & \textbf{0.3231} & 14.107         & \textbf{0.3235} & 13.694         & \textbf{0.3236}\\\hline\hline
\end{tabular}
\end{table*}
\begin{table*}[h]
\caption{\textbf{The trade-off curve of SD-v2.1}, where the best FID-30k and CLIP Score are highlighted.}\label{tab:sdv21}
\begin{tabular}{c|cc|cc|cc|cc} \hline\hline
         & \multicolumn{4}{c}{DDIM}                                              & \multicolumn{4}{c}{{ DPMSolver}++}                                       \\\cline{2-9}
         & \multicolumn{2}{c}{CFG}           & \multicolumn{2}{c}{S-CFG}         & \multicolumn{2}{c}{CFG}           & \multicolumn{2}{c}{S-CFG}         \\\hline
$\gamma$ & FID-30K         & CLIP Score      & FID-30K         & CLIP Score      & FID-30K         & CLIP Score      & FID-30K         & CLIP Score      \\\hline
2.0      & 14.394          & 0.3053          & 13.892          & 0.3068          & 14.999          & 0.3040          & 14.864          & 0.3060          \\
3.0      & 10.509          & 0.3191          & 10.227          & 0.3204          & 10.869          & 0.3187          & 10.797          & 0.3200          \\
5.0      & \textbf{10.429} & 0.3286          & \textbf{10.137} & 0.3306          & \textbf{10.241} & 0.3291          & \textbf{10.016} & 0.3304          \\
7.5      & 11.548          & 0.3331          & 11.278          & 0.3342          & 11.324          & 0.3339          & 10.944          & 0.3342          \\
10.0     & 12.604          & \textbf{0.3357} & 12.371          & \textbf{0.3359} & 12.166          & \textbf{0.3356} & 11.833          & \textbf{0.3359} \\\hline\hline
\end{tabular}
\end{table*}
\begin{table*}[h]
\caption{\textbf{The trade-off curve of IF}, where the best FID-30k and CLIP Score are highlighted.}\label{tab:sdv21}
\begin{tabular}{c|cc|cc|cc|cc}\hline\hline
         & \multicolumn{4}{c}{DDIM}                                           & \multicolumn{4}{c}{{ DPMSolver}++}                                     \\\cline{2-9}
         & \multicolumn{2}{c}{CFG}         & \multicolumn{2}{c}{S-CFG}        & \multicolumn{2}{c}{CFG}          & \multicolumn{2}{c}{S-CFG}        \\ \hline
$\gamma$ & FID-30K       & CLIP Score      & FID-30K        & CLIP Score      & FID-30K        & CLIP Score      & FID-30K        & CLIP Score      \\ \hline
2.0      & \textbf{9.820} & 0.3076          & \textbf{9.309} & 0.299           & \textbf{7.242} & 0.2997          & 8.494          & 0.2926          \\
3.0      & 13.804        & 0.3195          & 10.864         & 0.3152          & 7.799          & 0.3147          & \textbf{7.227} & 0.314           \\
5.0      & 17.267        & 0.3257          & 14.473         & 0.3259          & 11.396         & 0.3233          & 9.67           & 0.3226          \\
7.5      & 18.532        & 0.329           & 16.621         & 0.3288          & 13.968         & 0.327           & 12.402         & 0.3265          \\
10.0     & 19.029        & \textbf{0.3296} & 17.634         & \textbf{0.3299} & 15.31          & \textbf{0.3280} & 13.99          & \textbf{0.3280} \\\hline\hline
\end{tabular}
\end{table*}
\begin{table*}[h]
\caption{\textbf{The trade-off curve in the ablation 
 analysis }, where the best FID-30k and CLIP Score are highlighted. The experiment is based on SD-v1.5 with 50-step { DPMSolver}++ Sampler.}\label{tab:sdv21}
\begin{tabular}{c|cc|cc|cc|cc}\hline\hline
         & \multicolumn{2}{c}{S-CFG-mean}   & \multicolumn{2}{c}{S-CFG w/o sa} & \multicolumn{2}{c}{S-CFG-sa}     & \multicolumn{2}{c}{S-CFG}        \\\hline
$\gamma$ & FID-30K        & CLIP Score      & FID-30K        & CLIP Score      & FID-30K        & CLIP Score      & FID-30K        & CLIP Score      \\ \hline
2.0      & 10.703         & 0.2869          & 9.110          & 0.2963          & 9.063          & 0.2966          & 9.023          & 0.2964          \\
3.0      & \textbf{7.695} & 0.3044          & \textbf{7.811} & 0.3089          & \textbf{7.736} & 0.3099          & \textbf{7.717} & 0.3099          \\
5.0      & 8.813          & 0.3162          & 9.822          & 0.3185          & 9.755          & 0.3185          & 9.757          & 0.3187          \\
7.5      & 11.204         & 0.3213          & 12.102         & 0.3222          & 12.083         & 0.3227          & 12.059         & 0.3226          \\
10.0     & 12.838         & \textbf{0.3233} & 13.722         & \textbf{0.3235} & 13.690         & \textbf{0.3235} & 13.694         & \textbf{0.3236}\\\hline\hline
\end{tabular}
\end{table*}

\section{Additional Qualitative Samples}
\begin{figure*}
    \centering
    \includegraphics[width=1\linewidth]{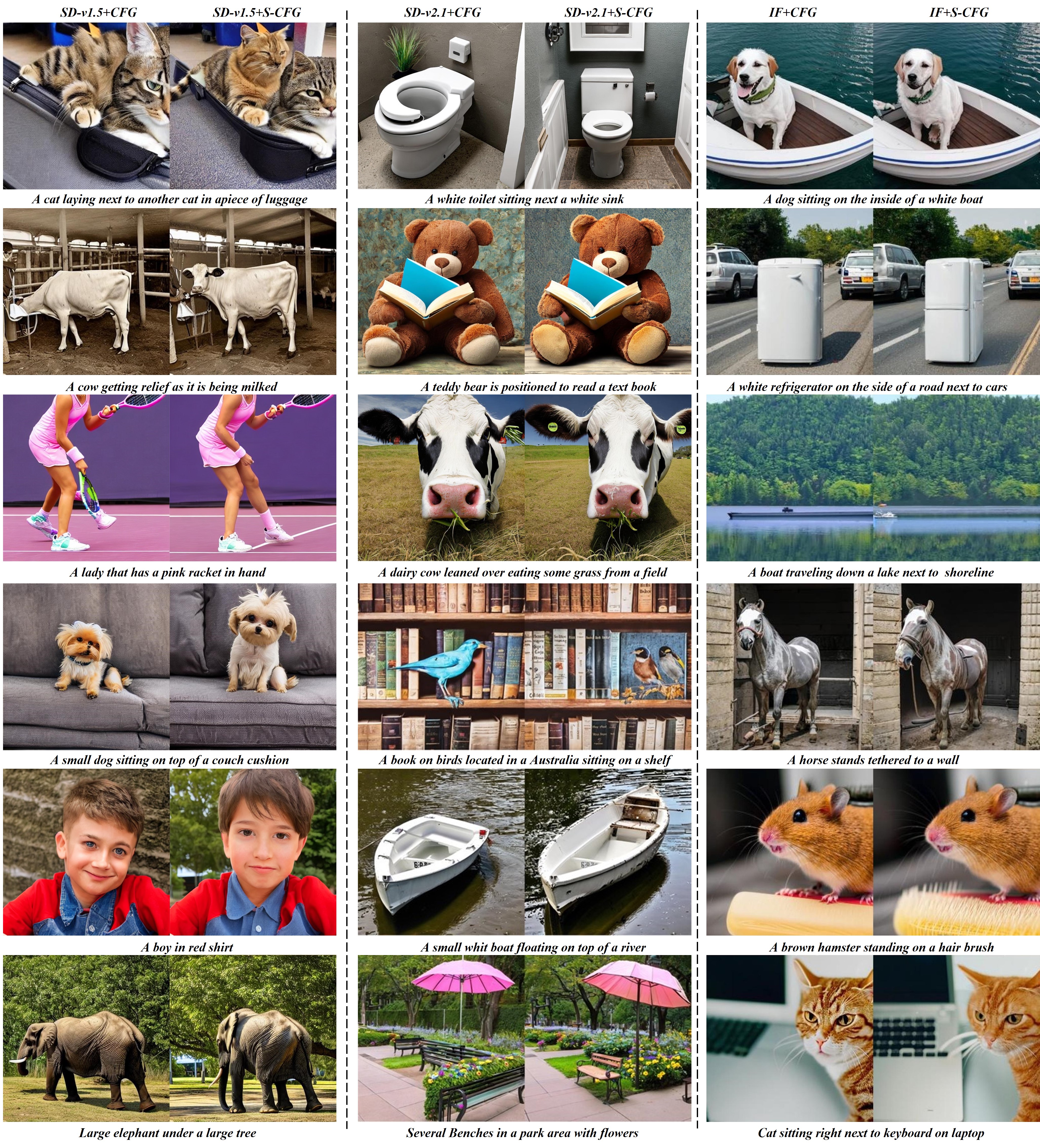}
    \caption{ \textbf{More samples generated by different base models with CFG (left) or S-CFG (right).}}
    \label{fig:suppcase}
\end{figure*}

In this section, we present supplementary samples in Figure~\ref{fig:suppcase} generated by different base models with CFG and S-CFG.
These additional samples further exhibit the superiority of S-CFG compared with the original CFG strategy.

% the comparisons of the inference time cost between the original CFG strategy and our S-CFG.

% \section{Rationale}
% \label{sec:rationale}
% % 
% Having the supplementary compiled together with the main paper means that:
% % 
% \begin{itemize}
% \item The supplementary can back-reference sections of the main paper, for example, we can refer to \cref{sec:intro};
% \item The main paper can forward reference sub-sections within the supplementary explicitly (e.g. referring to a particular experiment); 
% \item When submitted to arXiv, the supplementary will already included at the end of the paper.
% \end{itemize}
% % 
% To split the supplementary pages from the main paper, you can use \href{https://support.apple.com/en-ca/guide/preview/prvw11793/mac#:~:text=Delete%20a%20page%20from%20a,or%20choose%20Edit%20%3E%20Delete).}{Preview (on macOS)}, \href{https://www.adobe.com/acrobat/how-to/delete-pages-from-pdf.html#:~:text=Choose%20%E2%80%9CTools%E2%80%9D%20%3E%20%E2%80%9COrganize,or%20pages%20from%20the%20file.}{Adobe Acrobat} (on all OSs), as well as \href{https://superuser.com/questions/517986/is-it-possible-to-delete-some-pages-of-a-pdf-document}{command line tools}.